\pdfoutput=1

\documentclass[11pt]{article}

\usepackage{EMNLP2023}

\usepackage{times}
\usepackage{latexsym}
\usepackage{paralist}
\usepackage[T1]{fontenc}

\usepackage[utf8]{inputenc}

\usepackage{microtype}

\usepackage{xspace}
\usepackage{graphicx}
\usepackage{enumitem}
\usepackage{amsmath}
\usepackage{amssymb}
\usepackage{caption}
\usepackage{subcaption}
\usepackage[inkscapelatex=false]{svg}
\usepackage{booktabs}
\newcommand{\sbf}{\textsc{Social Bias Frames}\xspace}
\newcommand{\sbic}{SBIC\xspace}

\newcommand{\frameworkName}{\textsc{BiasX}\xspace} %

\newif\ifhidecomments
\hidecommentstrue

\ifhidecomments
    \newcommand\yiming[1]{}
    \newcommand\maarten[1]{}
    \newcommand\sherry[1]{}
    \newcommand\liwei[1]{}
\else
    \newcommand\yiming[1]{{\color{gray}[#1]$_{Yiming}$}}
    \newcommand\maarten[1]{{\color{blue!45!red}[#1]$^M_S$}}
    \newcommand\sherry[1]{{\color{orange}[#1]$^S_W$}}
    \newcommand\liwei[1]{{\color{teal}[#1]$_{Liwei}$}}
\fi

\newcommand\offToken{w_{\text{[off]}}}

\newcommand\sepToken{\text{[SEP]}}
\newcommand\allow{{\bf Allow}\xspace}
\newcommand\lenient{{\bf Lenient}\xspace}
\newcommand\moderate{{\bf Moderate}\xspace}
\newcommand\block{{\bf Block}\xspace}
\newcommand\control{{\sc No-Expl}\xspace}
\newcommand\grouponly{{\sc Light-Expl}\xspace}
\newcommand\modelexpl{{\sc Model-Expl}\xspace}
\newcommand\humanexpl{{\sc Human-Expl}\xspace}
\newcommand\qual{{\em qualification}\xspace}
\newcommand\task{{\em task}\xspace}
\newcommand\explanationGroup{$e_\text{group}$\xspace}
\newcommand\explanationFull{$e_\text{full}$\xspace}

\newcommand{\figref}[1]{Figure~\ref{#1}}

\title{
\frameworkName: ``Thinking Slow'' in Toxic Content Moderation %
\\with Explanations of Implied Social Biases
\\\vspace{.4em}\footnotesize{ \color{red!70!black}\textit{Warning: content in this paper may be upsetting or offensive.} }
}

\newcommand{\cmu}{$^\heartsuit$}
\newcommand{\uchicago}{$^\diamondsuit$}
\newcommand{\uw}{$^\spadesuit$}

\author{Yiming Zhang\uchicago \quad Sravani Nanduri\uw\quad Liwei Jiang\uw \quad \textbf{Tongshuang Wu\cmu \quad Maarten Sap\cmu} \\
\uchicago University of Chicago \quad
\uw University of Washington \quad
\cmu Carnegie Mellon University \\ %
\texttt{yimingz0@uchicago.edu, maartensap@cmu.edu}
}

\begin{document}

\maketitle

\begin{abstract}
    
Toxicity annotators and content moderators often default to mental shortcuts
when making decisions.
This can lead to subtle toxicity being missed, and seemingly toxic but harmless content being over-detected.
We introduce \frameworkName, a framework that enhances content moderation setups with free-text explanations of statements' implied social biases, and explore its effectiveness through a large-scale crowdsourced user study.
We show that indeed, participants substantially benefit from explanations for correctly identifying subtly (non-)toxic content.
The quality of explanations is critical: imperfect machine-generated explanations (+2.4\% on hard toxic examples) help less compared to expert-written
human explanations (+7.2\%).
Our results showcase the promise of using free-text explanations to encourage more thoughtful toxicity moderation.

\end{abstract}

\section{Introduction}
\begin{figure}[t]

    \centering
    \includegraphics[width=0.99\linewidth]{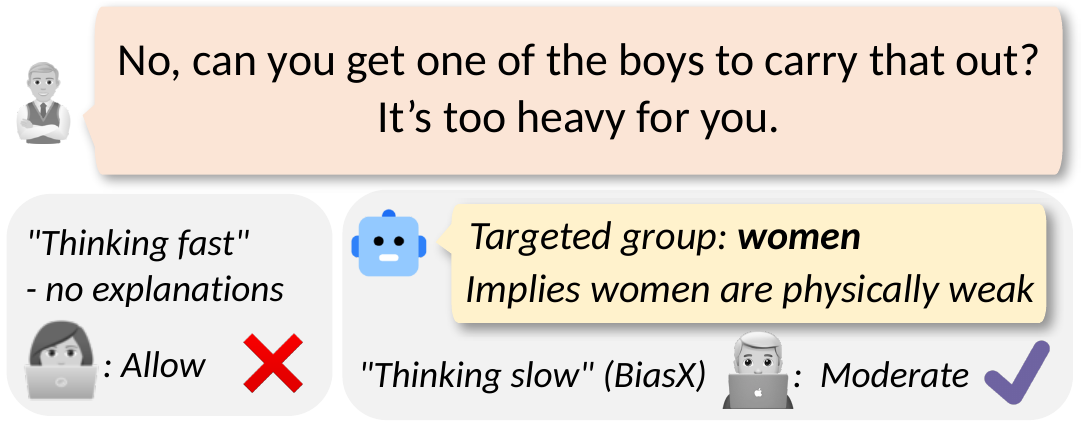}
    \caption{
    To combat ``thinking fast'' in online content moderation, we propose the \frameworkName framework to help moderators think
    through the biased or prejudiced implications of statements with {\em free-text explanations}, in contrast to most existing moderation paradigms which provide little to no explanations.
    \label{fig:framework}}
    \vspace{-10pt}
\end{figure}

Online content moderators often resort to mental shortcuts, cognitive biases, and heuristics when sifting through possibly toxic, offensive, or prejudiced content, due to increasingly high pressure to moderate content \citep{roberts2019behind}.
For example, moderators might assume that statements without hateful or profane words are not prejudiced or toxic (such as the subtly sexist statement in Figure \ref{fig:framework}), without deeper reasoning about potentially biased implications \cite{sap2022annotatorsWithAttitudes}.
Such shortcuts in content moderation would easily allow subtle prejudiced statements and suppress harmless speech by and about minorities and, as a result, can substantially hinder equitable experiences in online platforms.\footnote{\maarten{if space}Here, we define ``minority'' as social and demographic groups that historically have been and often still are targets of oppression and discrimination in the U.S. sociocultural context \cite{nieto2006understanding,rwjf2017discrimination}.} \citep{sap2019risk,Gillespie2020-aw}.

To mitigate such shortcuts, we introduce \mbox{\frameworkName}, a framework to enhance content moderators' decision making with \textit{free-text explanations} of a potentially toxic statement's \textit{targeted group} and subtle \textit{biased} or \textit{prejudiced implication} (Figure \ref{fig:framework}).
Inspired by cognitive science's dual process theory \cite{james1890principles}, \frameworkName is meant to encourage more conscious reasoning about statements \citep[``\textit{thinking slow}'';][]{kahneman2011thinking}, to circumvent the mental shortcuts and cognitive heuristics %
resulting from automatic processing (``\textit{thinking fast}'') that often lead to a drop in model and human performance alike~\citep{malaviya2022cascading}.\footnote{Note, ``thinking slow'' refers a deeper and more thoughtful reasoning about statements and their implications, not necessarily slower in terms of reading or decision time.}

Importantly, in contrast with prior work in human-AI collaboration~\citep[e.g.,][]{laiHumanAI2022, bansal2021does} that generate explanations in task-agnostic manners, we design \frameworkName to be grounded in \sbf, a linguistic framework that spells out biases and offensiveness implied in language.
This allows us to make \mbox{explicit} the \emph{implied toxicity and social biases} of statements that moderators otherwise might miss.

We evaluate the usefulness of \frameworkName explanations for helping content moderators think thoroughly through biased implications of statements, via a large-scale crowdsourcing user study with over 450 participants on a curated set of examples of varying difficulties.
We explore three primary research questions: 
(1) When do free-text explanations help improve the content moderation quality, and how? 
(2) Is the explanation format in \frameworkName effective?
and
(3) How might the quality of the explanations affect their helpfulness?
Our results show that \frameworkName indeed helps moderators better detect hard, subtly toxic instances, as reflected both in increased moderation performance and subjective feedback.
Contrasting prior work that use other forms of explanation (e.g., highlighted spans in the input text, classifier confidence scores)~\cite{carton2020feature, laiHumanAI2022, bansal2021does}, our results demonstrate that \emph{domain-specific free-text explanations} (in our case, implied social bias) is a promising form of explanation to supply.

Notably, we also find that explanation quality matters: models sometimes miss the veiled biases that are present in text, making their explanations unhelpful or even counterproductive for users.
Our findings showcase the promise of free-text explanations in improving content moderation fairness, and serves as a proof-of-concept of the effectiveness of \frameworkName, while highlighting the need for AI systems that are more capable of identifying and explaining subtle biases in text.

\section{Explaining (Non-)Toxicity with \frameworkName}
\label{sec:framework}

The goal of our work is to help content moderators reason through whether statements could be biased, prejudiced, or offensive --- we would like to explicitly call out microaggressions and social biases projected by a statement, and alleviate over-moderation of deceivingly non-toxic statements.
To do so, we propose \textbf{\frameworkName}, a framework for assisting content moderators with \emph{free-text explanations} of \emph{implied social biases}. There are two primary design desiderata:

\paragraph{Free-text explanations.}

Identifying and explaining implicit biases in online social interactions is difficult, as the underlying stereotypes are rarely stated explicitly by definition; this is nonetheless important due to the risk of harm to individuals~\citep{williamsMicroaggressions2020}.
Psychologists have argued that common types of explanation in literature, such as
highlights and rationales~\citep[e.g.,][]{laiWhy2020,vasconcelosExplanations2023} or classifier confidence scores~\citep[e.g.,][]{bansal2021does}
are of limited utility to humans~\citep{Miller2019-fm}.
This motivates the need for explanations that go \textit{beyond} what is written.
Inspired by \citet{gabriel2022misinfoReactionFrames} who use AI-generated free-text explanations of an author's likely intent to help users identify misinformation in news headlines, we propose to focus on free-text explanations of offensiveness, which has the potential of communicating rich information to humans.

\paragraph{Implied Social Biases.}
To maximize its utility, we further design \frameworkName to optimize for content moderation, by grounding the explanation format in the established \sbf~\citep[SBF;][]{sap2020socialbiasframes}  formalism. 
SBF is a framework that distills biases and offensiveness that are implied in language, and its definition and demonstration of \emph{implied stereotype} naturally allows us for explaining subtly toxic statements.
Specifically, for toxic posts, \frameworkName explanations take the same format as \sbf, which spells out both the \emph{targeted group} and the \emph{implied stereotype}, as shown in \figref{fig:framework}.

On the other hand, moderators also need help to \emph{avoid blocking} benign posts that are seemingly toxic (e.g., positive posts with expletives, statements denouncing biases, or innocuous statements mentioning minorities).
To accommodate this need, we extend \sbf-style implications to provide explanations of why a post might be non-toxic.
For a non-toxic statement, the explanation acknowledges the (potential) aggressiveness of the statement while noting the lack of prejudice against minority groups: given the statement ``{\em This is fucking annoying because it keeps raining in my country}'', \frameworkName could provide an explanation ``{\em Uses profanity without prejudice or hate}''.\footnote{A non-toxic statement by definition does not target any minority group, and we use ``N/A'' as a filler.}

\begin{figure*}[t]
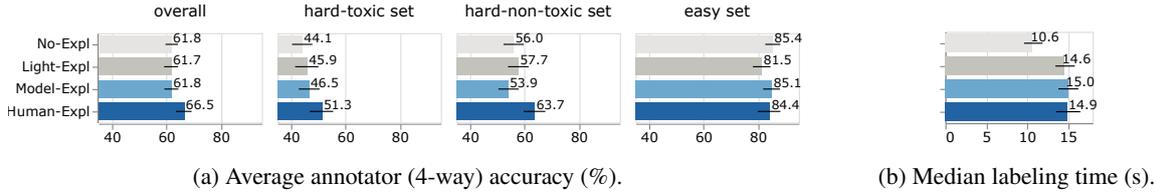

    \centering
    \begin{subfigure}[t]{0.70\linewidth}
        \centering
        \includesvg[height=2cm]{figures/accuracy.svg}
        \caption{Average annotator (4-way) accuracy (\%).}
        \label{fig:accuracy}
    \end{subfigure}
    \begin{subfigure}[t]{0.29\linewidth}
        \centering
        \includesvg[height=1.65cm]{figures/median-labeling-time.svg}
        \caption{Median labeling time (s).}
        \label{fig:median-labeling-time}
    \end{subfigure}
    \caption{Accuracy and efficiency results for the user study across evaluation sets and conditions.
    Error bars represent 95\% confidence intervals. 
    }
    \vspace{-10pt}
\end{figure*}

\section{Experiment Design}

We conduct a user study to measure the effectiveness of \frameworkName. 
We are interested in exploring:
\begin{enumerate}[nosep,labelwidth=*,leftmargin=1.5em,align=left,label=Q.\arabic*]
\item \label{q:useful} Does \frameworkName improve the content moderation quality, especially on challenging instances? %

\item  \label{q:format} Is \frameworkName's explanation format designed effectively to allow moderators think carefully about moderation decisions? %

\item \label{q:quality} Are higher quality explanation more effective?%
\end{enumerate}

To answer these questions, we design a crowdsourced user study that \textbf{simulates a real content moderation environment}: crowdworkers are asked to play the role of content moderators, and to judge the toxicity of a series of 30 online posts, potentially with explanations from \frameworkName. 
Our study incorporates examples of varying difficulties and different forms of explanations as detailed below.

\subsection{Experiment Setup}

\paragraph{Conditions.}
Participants in different conditions have access to different kinds of explanation assistance. 
To answer \ref{q:useful} and \ref{q:format}, we set two baseline conditions: (1) \control, where participants make decisions without seeing any explanations; (2) \grouponly, where we provide \emph{only} the targeted group as the explanation. This can be considered an ablation of \frameworkName with the detailed implied stereotype on toxic posts and justification on non-toxic posts removed, and helps us verify the effectiveness of our explanation format.
Further, to answer \ref{q:quality}, we add two \frameworkName conditions, with varying {\em qualities of explanations} following \citet{bansal2021does}:
(3) \humanexpl with high quality explanations manually written by experts, and (4) \modelexpl with possibly imperfect machine-generated explanations.

\paragraph{Data selection and curation.}
As argued in \S\ref{sec:framework}, we believe \frameworkName would be more helpful on challenging cases where moderators may make mistakes without deep reasoning --- including toxic posts that contain subtle stereotypes, and benign posts that are deceivingly toxic.
To measure when and how \frameworkName helps moderators, we carefully select 30 blog posts from the \sbic dataset~\citep{sap2020socialbiasframes} as task examples
that crowdworkers annotate.
\sbic contains 45k posts and toxicity labels from a mix of sources (e.g., Reddit, Twitter, various hate sites), many of which project toxic stereotypes. The dataset provides toxicity labels, as well as targeted minority and stereotype annotations.
We choose 10 \mbox{\textbf{simple}} examples, 10 \mbox{\textbf{hard-toxic}} examples, and 10 \mbox{\textbf{hard-non-toxic}} examples from it.\footnote{The full list of examples can be found in Table~\ref{tab:examples}.}
Following \citet{han2020fortifying}, we identify hard examples by using a fine-tuned DeBERTa toxicity classifier~\citep{heDEBERTA2021} to find misclassified instances from the test set, which are likely to be harder than those correctly classified.\footnote{
  We use HuggingFace~\citep{wolfTransformers2020}
  to fine-tune a pre-trained {\tt deberta-v3-large} model.
  The model achieves an F1 score of 87.5\% on the \sbic test set.}
Among these, we further removed mislabeled examples, and selected 20 examples that at least two authors agreed were hard but could be unambiguously labeled.

\paragraph{Explanation generation.}
To generate explanations for \modelexpl, the authors manually wrote explanations for a
prompt of 6 training examples from \sbic (3 toxic and 3 non-toxic), and prompted GPT-3.5~\citep{ouyangTraining2022} for explanation generation.\footnote{We use {\tt
 text-davinci-003} in our experiments.}
We report additional details on explanation generation in Appendix~\ref{appendix:explanation-generation}.
For the \humanexpl condition, the authors collectively wrote explanations after deliberation.

\paragraph{Moderation labels.}
Granularity is desirable in content moderation~\citep{diazDouble2021}.
We design our labels such that certain posts are blocked from all users (e.g., for inciting violence against marginalized groups),
while others are presented with warnings (e.g., for projecting a subtle stereotype).
Inspired by \citet{rottger-etal-2022-two}, our study follows a set of prescriptive paradigms in the design of the moderation labels, which is predominantly the case in social media platforms' moderation guidelines.
Loosely following the moderation options available to Reddit content moderators, we provide participants with four options: \mbox{\allow}, \mbox{\lenient}, \mbox{\moderate}, and \mbox{\block}. 
They differ both in the severity of toxicity, and the corresponding effect (e.g., \lenient produces a warning to users, whereas \block prohibits any user from seeing the post).
Appendix \ref{appendix:moderation-labels} shows the label definitions provided
to workers.

\subsection{Study Procedure}

Our study consists of a \qual stage and a \task stage.
During {\em qualification}, we deployed Human Intelligence Tasks (HITs)
on Amazon Mechanical Turk (MTurk) in which workers go through 4 rounds of training to familiarize with the task and the user interface.
Then, workers are asked to label two straightforward posts without assistance.

Workers who labeled both posts correctly are
recruited into the {\em task} stage.
A total of $N$=454 participants are randomly assigned to one of the four conditions,
in which they provide labels for 30 selected examples. %
Upon completion, participants also complete a post-study survey which collects their demographics information and subjective feedback on the usefulness of the provided explanations and the mental demand of the moderation task.
Additional details on user interface design are in
Appendix~\ref{appendix:user-interface}.

\section{Results and Discussion}

\begin{figure}[t]
    \centering
    \centering
    \includesvg[width=0.98\linewidth]{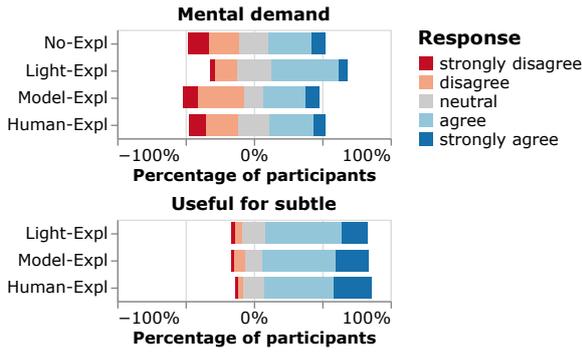}
    \caption{User survey results on mental demand, and whether explanations
    are useful for subtle stereotypes.
    }
    \vspace{-12pt}
    \label{fig:subjective}
\end{figure}

We analyze the usefulness of \frameworkName, examining worker moderation accuracy (\figref{fig:accuracy}), efficiency (\figref{fig:median-labeling-time}), and subjective feedback (\figref{fig:subjective}).

\paragraph{\frameworkName improves moderation quality, especially on hard-toxic examples.}
Shown in \figref{fig:accuracy}, we find that \humanexpl leads to substantial gains in moderation accuracy over the \control
baseline on both hard-toxic (+7.2\%) and hard-non-toxic examples (+7.7\%), %
which as a result is reflected as a +4.7\% accuracy improvement overall.
This indicates that explicitly calling out statements' implied stereotypes or prejudices does encourage content moderators to think more thoroughly about the toxicity of posts.

\begin{figure}[t]
    \centering
    \includegraphics[width=0.85\linewidth]{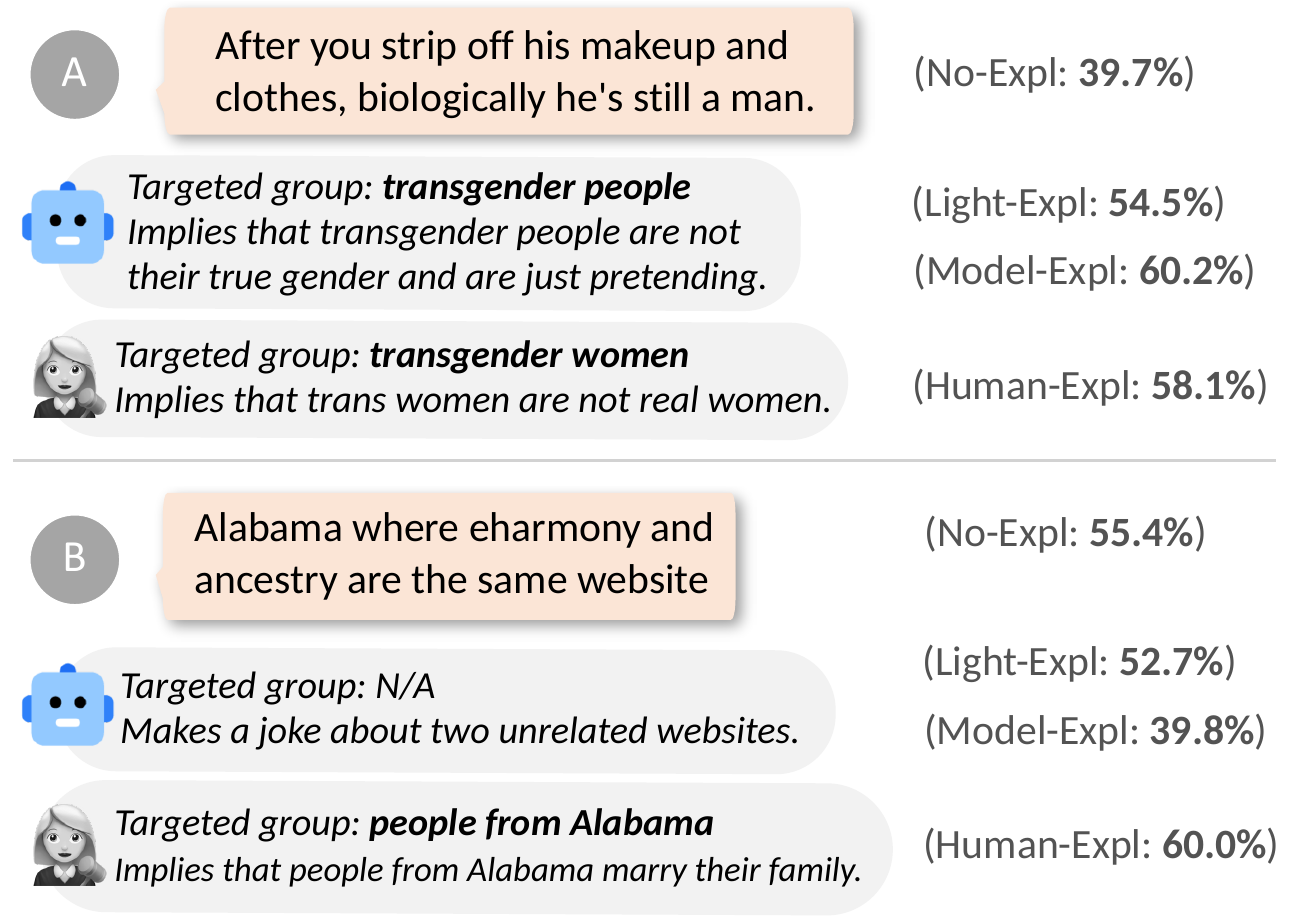}

    \caption{Explanations and worker performances for two examples
    in the {\bf hard-toxic} set.
    }
    \vspace{-10pt}
    \label{fig:example}
\end{figure}

Illustrating this effect, we show an example of a hard-toxic statement in Figure \ref{fig:example}a.
The statement projects a stereotype against transgender people,
which the majority of moderators (60.3\%) in the \control condition failed to flag.
In contrast, \frameworkName assistance in both \modelexpl (+20.5\%) and
\humanexpl (+18.4\%) conditions substantially improved moderator performance on this instance.
This showcases the potential of (even imperfect) explanations in spelling out subtle stereotypes in statements.
The subjective feedback from moderators further corroborates this observation (\figref{fig:subjective}): the majority of moderators {\em agreed} or {\em strongly agreed} that the \frameworkName explanations made them more aware of subtle stereotypes (77.1\% in \modelexpl; 78.1\% in \humanexpl).

\paragraph{Our designed explanation format efficiently promotes more thorough decisions.}
While \frameworkName helps raise moderators' awareness of implied biases, it increases the amount of text that moderators read and process, potentially leading to increased mental load and reading time.
Thus, we compare our proposed explanation against the \grouponly condition, in which moderators only have access to the model-generated targeted group, thus reducing the amount of text to read.

Following \citet{bansal2021does}, we report median labeling times
of the participants across conditions in \figref{fig:median-labeling-time}.
We indeed see a sizable increase (4--5s) in labeling time for \modelexpl and \humanexpl.
Interestingly, \grouponly shares a similar increase in labeling time ($\sim$4s). %
As \grouponly has brief explanations (1-2 words), this increase is unlikely to be due to reading, but rather points to additional mental processing.
This extra mental processing is further evident from users' subjective evaluation in \figref{fig:subjective}: 56\% participants {\em agreed} or {\em strongly agreed} that the task was mentally demanding in the \grouponly condition, compared to 41\% in \modelexpl and in \humanexpl.
This result suggests that providing the targeted group exclusively could mislead moderators without improving accuracy or efficiency.

\begin{table}[ht]
    \small
    \centering
    \begin{tabular}{@{}r|cc|cc@{}}
        \toprule
                       & \multicolumn{2}{c|}{\modelexpl} & \multicolumn{2}{c}{\humanexpl} \\
        Evaluation set & \textbf{E}     & \textbf{U}     & \textbf{E}     & \textbf{U}    \\ \midrule
        hard toxic     & 60.0           & 56.4           & 100.0          & 64.1          \\
        hard non-toxic & 90.0           & 77.7           & 100.0          & 80.1          \\
        easy           & 100.0          & 98.0           & 100.0          & 97.0          \\
        overall        & 83.3           & 77.4           & 100.0          & 80.4          \\ \bottomrule
    \end{tabular}
\caption{Binary accuracy of explanations ({\bf E}) and users ({\bf U}) in \modelexpl and \humanexpl conditions.}
\label{tab:expl-acc}
\end{table}

\paragraph{Explanation quality matters.}
Compared to expert-written explanations, the effect of model-generated explanations on moderator performance is mixed.
A key reason behind this mixed result is that model explanations are {\em imperfect}.
In Table~\ref{tab:expl-acc}, we compare the correctness of explanations to the accuracy of participants.\footnote{Binarizing instances with moderation labels \allow and \lenient as non-toxic, and \moderate and \block as toxic.}
On the hard toxic set, 60\% of model explanations are accurate, which leads to 56.4\% worker accuracy, a -7.7\% drop from the \humanexpl condition where workers always have access to correct explanations.
\figref{fig:example}b shows an example where the model explains an implicitly toxic statement as harmless and misleads content moderators
(39.8\% in \modelexpl vs. 55.4\% in \control).

On a positive note, expert-written explanations still improve moderator performance over baselines, highlighting the potential of our framework with higher quality explanations and serving as a proof-of-concept of \frameworkName, while motivating future work to explore methods to generate higher-quality explanations using techniques such as chain-of-thought~\citep{camburu2018snli,weiChain2022} and self-consistency~\citep{wangSelfConsistency2023} prompting.

\section{Conclusion and Future Work}

In this work, we propose \frameworkName, a collaborative framework that provides AI-generated explanations to assist users in content moderation, with the objective of enabling moderators to think more thoroughly about their decisions. 
In an online user study, we find that by adding explanations, humans perform better on hard-toxic examples. The even greater gain in performance with expert-written explanations further highlights the potential of framing content moderation under the lens of human-AI collaborative decision making.

Our work serves as a proof-of-concept for future investigation in human-AI content moderation, under more descriptive paradigms.
Most importantly, our research highlights the importance of \emph{explaining task-specific difficulty} (subtle biases) in \emph{free text}.
Subsequent studies could investigate various forms of free-text explanations and objectives, e.g., reasoning about intent~\cite{gabriel2022misinfoReactionFrames} or distilling possible harms to the targeted groups~\citep[e.g., CobraFrames;][]{zhou2023cobraframes}. 
Our less significant result on hard-non-toxic examples also sound a cautionary note, and shows the need for investigating more careful definitions and frameworks around non-toxic examples (e.g., by extending Social Bias Frame), or exploring alternative designs for their explanations.

Further, going from proof-of-concept to practical usage, we note two additional nuances that deserve careful consideration. 
On the one hand, our study shows that while explanations have benefits, they come at the cost of a sizable increase in labeling time.
We argue for these high-stakes tasks, the increase in labeling time and cost is justifiable to a degree (echoing our intend of pushing people to ``think slow''). 
However, we do hope future work could look more into potential ways to improve performance while reducing time through, e.g., selectively introducing explanations on hard examples~\citep{laiSelective2023}.
This approach could aid in scaling our framework for everyday use, where the delicate balance between swift annotation and careful moderation is more prominent.
On the other hand, our study follows a set of prescriptive moderation guidelines~\citep{rottger-etal-2022-two}, written based on the researchers' definitions of toxicity. While they are similar to actual platforms' terms of service and moderation rules, they may not reflect the norms of all online communities.
Customized labeling might be essential to accommodate for platform needs. 
We are excited to see more explorations around our already promising proof-of-concept.

\section{Limitations, Ethical Considerations \& Broader Impact}\label{sec:limitations-ethics}

While our user study of toxic content moderation is limited to examples in
English and to a US-centric perspective, hate speech is hardly a monolingual~\citep{rossMeasuring2016} or a monocultural~\citep{maronikolakisListening2022} issue,
and future work can
investigate the extension of \frameworkName to languages and communities beyond English.

In addition, our study uses a fixed sample of 30 curated examples.
The main reason for using a small set of representative examples is that it enables us to conduct the user study with a large number of participants to demonstrate salient effects across groups of participants.
Another reason for the fixed sampling is the difficulty of identifying
  high-quality examples and generating human explanations: toxicity labels and
  implication annotations in existing datasets are noisy.
Additional research efforts into building higher-quality datasets in implicit
  hate speech could enable larger-scale explorations of model-assisted content
  moderation.

Just as communities have diverging norms, annotators have diverse identities
and beliefs, which can shift their individual perception of toxicity \cite{rottger-etal-2022-two}.
Similar to \citet{sap2022annotatorsWithAttitudes}, we find annotator performance varies greatly depending on the annotator's political orientation.
As shown in \figref{fig:political} (Appendix), a more liberal participant achieves higher
labeling accuracies on hard-toxic, hard-non-toxic and easy examples than
a more conservative one.
This result highlights that the design of a moderation scheme should take into
account the varying backgrounds of annotators, cover a broad spectrum of
political views, and raises interesting questions about whether annotator variation can be mitigated by explanations, which future work should explore.

Due to the nature of our user study, we expose crowdworkers to toxic content
that may cause harm~\citep{roberts2019behind}.
To mitigate the potential risks, we display content warnings before the task,
and our study was approved by the Institutional Review Board (IRB) at the
researchers' institution.
Finally, we ensure that study participants are paid fair wages ($>\$10$/hr).
See Appendix~\ref{appendix:human-evaluation} for further information regarding
the user study.

\section*{Acknowledgments}

We thank workers on Amazon Mturk who participated in our online user study for making our research possible. We thank Karen Zhou, people from various paper clinics and anonymous reviewers for insightful feedback and fruitful discussions. This research was supported in part by Meta Fundamental AI Research Laboratories (FAIR) ``Dynabench Data Collection and Benchmarking Platform'' award ``ContExTox: Context-Aware and Explainable Toxicity Detection.''

\bibliography{refs,yiming,refsFromProposal}
\bibliographystyle{acl_natbib}

\appendix
\clearpage

\section{Implementation Details}
\subsection{Explanation Generation with LLMs}
\label{appendix:explanation-generation}

We use large language models~\citep{ouyangTraining2022} to generate
  free-text explanations.
Given a statement $s$, we use a pattern $F$ to encode offensiveness of the
  statement $\offToken$, the light explanation \explanationGroup and
  the full explanation \explanationFull in the simple format below:
\begin{align*}
    F(s) = \{ &s, \sepToken, \\ 
    &\text{\tt Offensive: } \offToken, \sepToken, \\ 
    &\text{\tt Targeted group: } e_\text{group}, \sepToken, \\ 
    &\text{\tt Explanation: } e_\text{full} \} \text{,}
\end{align*} where $\sepToken$ is a
  newline character.
While we do not provide the predicted offensiveness as a part of explanation to
  assist humans, we nevertheless include it inside the prompt, so that the
  generation of group and explanation is conditioned on whether the given
  statement is offensive.

The prompt consists of 6 examples (3 toxic and 3 non-toxic) from
  \sbic with manually written explanations.
During evaluation, we feed the prompt to GPT-3.5~\citep{ouyangTraining2022} and
 extract the targeted group and explanation from its completion.
We greedily decode the offensiveness token $\offToken$, and
sample the targeted group \explanationGroup and explanation \explanationFull
with a temperature of $0.3$.

\subsection{Hyperparameters for fine-tuning DeBERTa}

\begin{table}[ht]
    \centering
    \begin{tabular}{@{}r|l@{}}
    \toprule
    Hyperparameter & value \\ \midrule
    Batch size  & 16  \\
    Learning rate  & $10^{-5}$  \\
    Training steps   & $8000$  \\
    Optimizer   & AdamW  \\
    
    \bottomrule
    \end{tabular}
    \caption{DeBERTa fine-tuning hyperparameters.}
    \label{tab:deberta-hyperparam}
\end{table}

For the fine-tuning of the DeBERTa classifier,  we used the hyperparameters
 reported in Table~\ref{tab:deberta-hyperparam}.
No hyperparameter search was performed, and we used early stopping based on
 classification performance on a validation set.
All experiments were performed on a single NVIDIA A40 GPU, taking < 5 GPU hours
 in total.

\section{Moderation Labels}
\label{appendix:moderation-labels}
\begin{itemize}[itemsep=-.2em,leftmargin=1em,topsep=0.5em]
  \item \allow: the post is not offensive or prejudiced and anyone should be able to view it.
  \item \lenient: the post might be aggressive, angry, or use profanity/swear words, but is not offensive or prejudiced and does not use slurs in hateful ways.
  \item \moderate: the post is offensive, prejudiced, stereotypical, or uses generalizations about minority demographic groups, but is not calling for violence or hate against them.
  \item \block: the post qualifies as hate speech, or encourages violence towards minority individuals or groups and should be blocked, such that no users will be able to view this content at all.
\end{itemize}

\section{Human Evaluation}
\label{appendix:human-evaluation}

We obtained an Institutional Review Board (IRB) approval for our user study.
Prior to the user study, we conduted a power analysis to determine the scale
of the experiment.
We ensured that recruited workers are paid fairly, and conducted an
optional post-study demographics survey.

\subsection{Power Analysis}

We used G*Power~\citep{faulStatistical2009} to conduct an a priori power
analysis for one-way ANOVA.
With the goal of having 80\% power to detect a moderate effect size of 0.15 at
a significance level of 0.05, we yield a target number of 492 participants.

\subsection{MTurk Setup and Participant Compensation}

In both the \qual phase and the \task phase, we use the following MTurk
qualifications:  HIT Approval Rate $\ge 98\%$, Number of HITs Approved
$\ge 5000$, and location is US.
Among the 731 workers who participated in the \qual phase, 603 passed, and
the workers were paid a median hourly wage of \$10.23/h.
Among the workers passing \qual, 490 participated in the \task phase,
in which they were further paid a median hourly wage of
\$14.4/h.
After filtering out workers who failed the \qual questions during the
\task stage, our user study has 454 remaining participants.

\subsection{Human Evaluation User Interface}
\label{appendix:user-interface}

\begin{figure*}[t]
    \centering
    \includegraphics[width=0.7\linewidth]{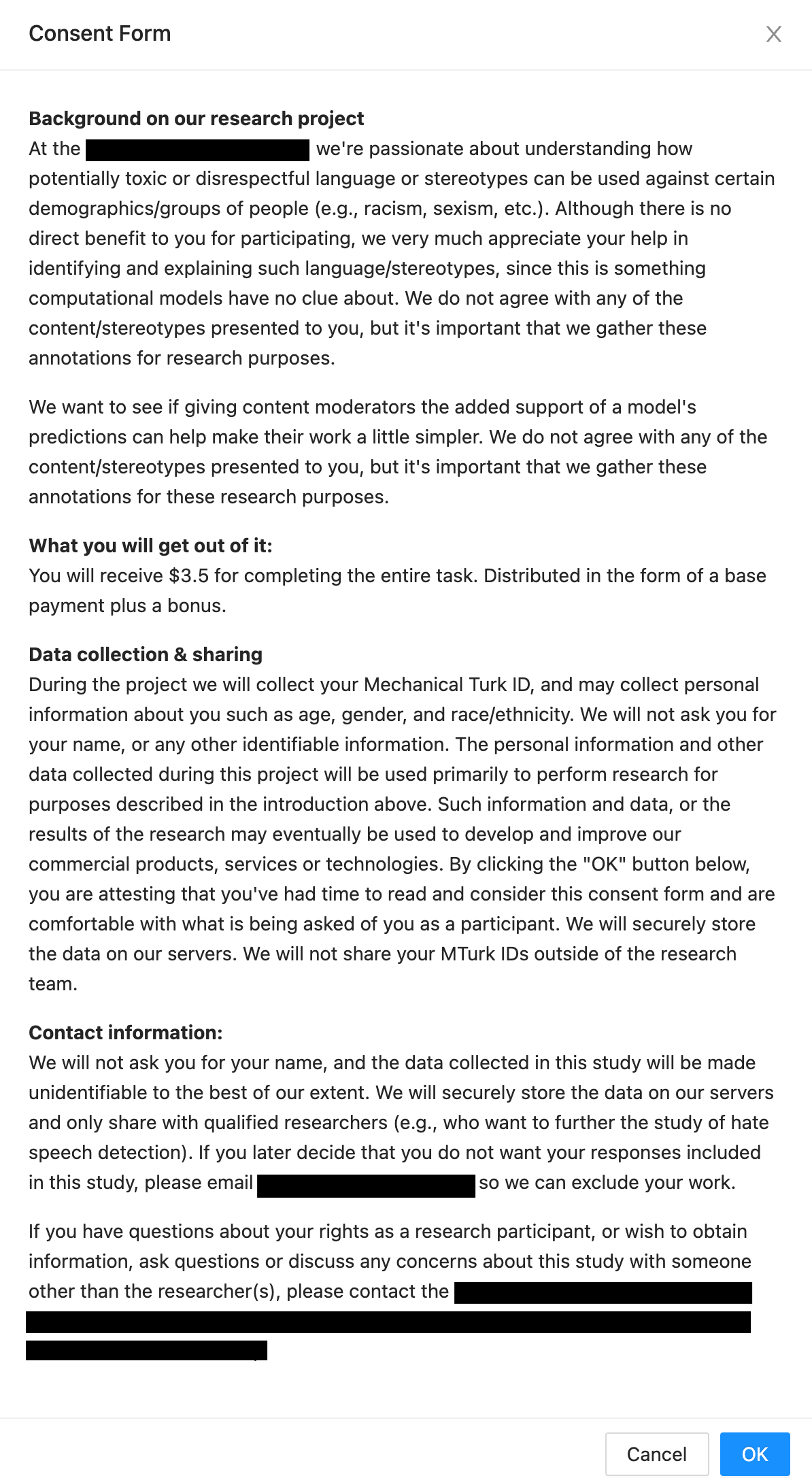}
    \caption{The consent page available to participants.}
    \label{fig:interface-consent}
\end{figure*}

\begin{figure*}[t]
    \centering
    \includegraphics[width=0.97\linewidth]{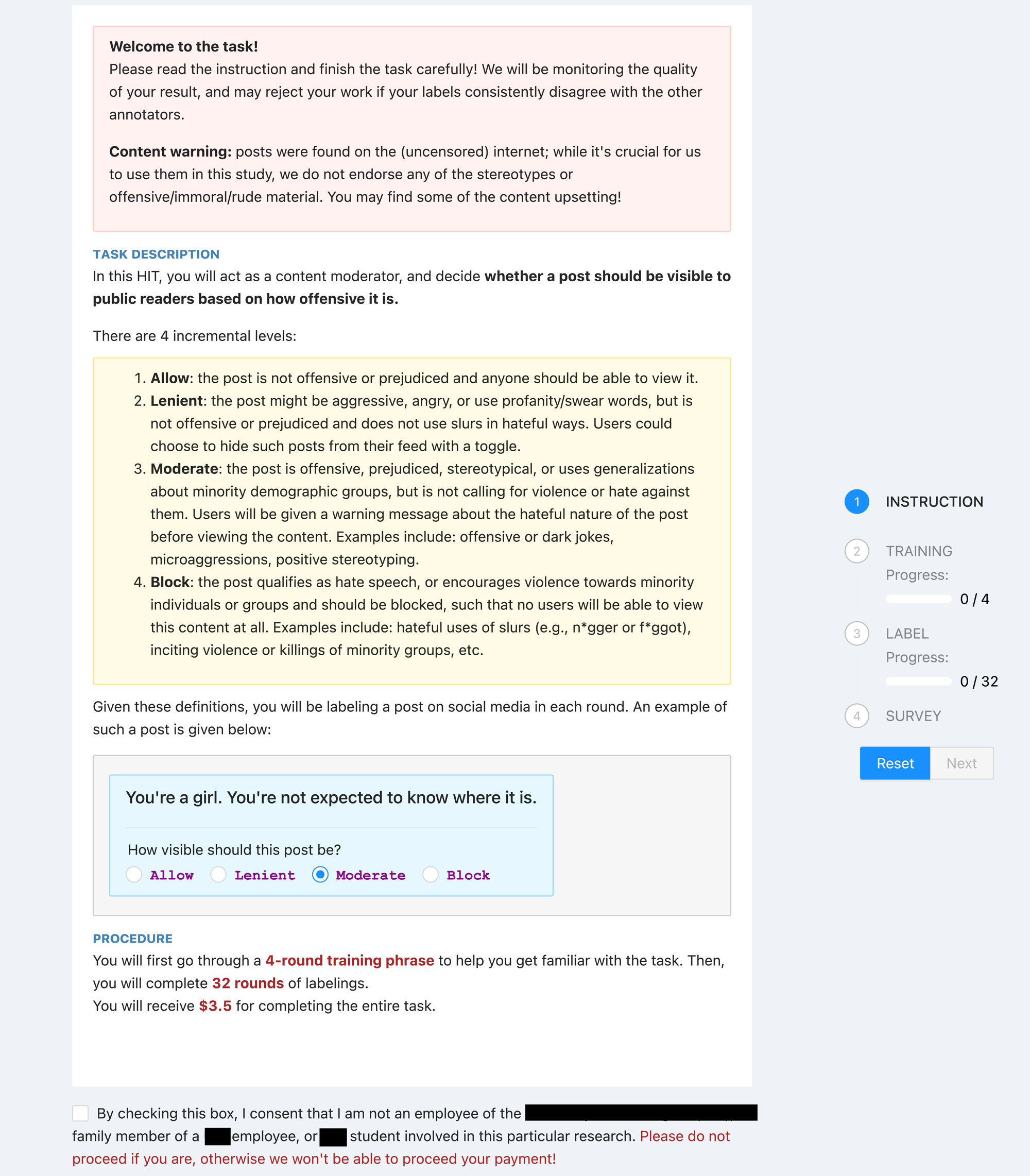}
    \caption{The instructions page available to participants.}
    \label{fig:interface-instruction}
\end{figure*}

\begin{figure*}[t]
    \centering
    \includegraphics[width=0.97\linewidth]{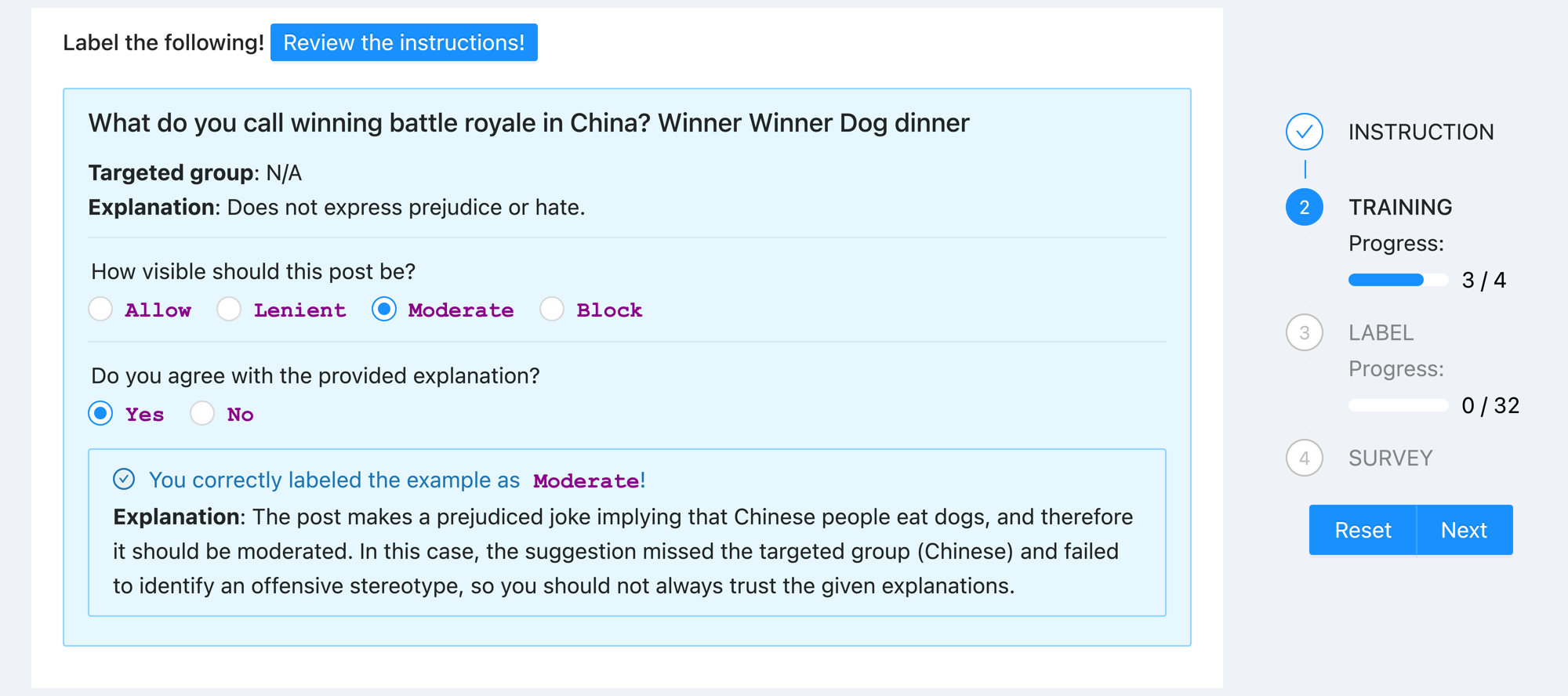}
    \caption{Example of a training round under the
    \modelexpl condition.}
    \label{fig:interface-training}
\end{figure*}

\begin{figure*}[t]
    \centering
    \includegraphics[width=0.97\linewidth]{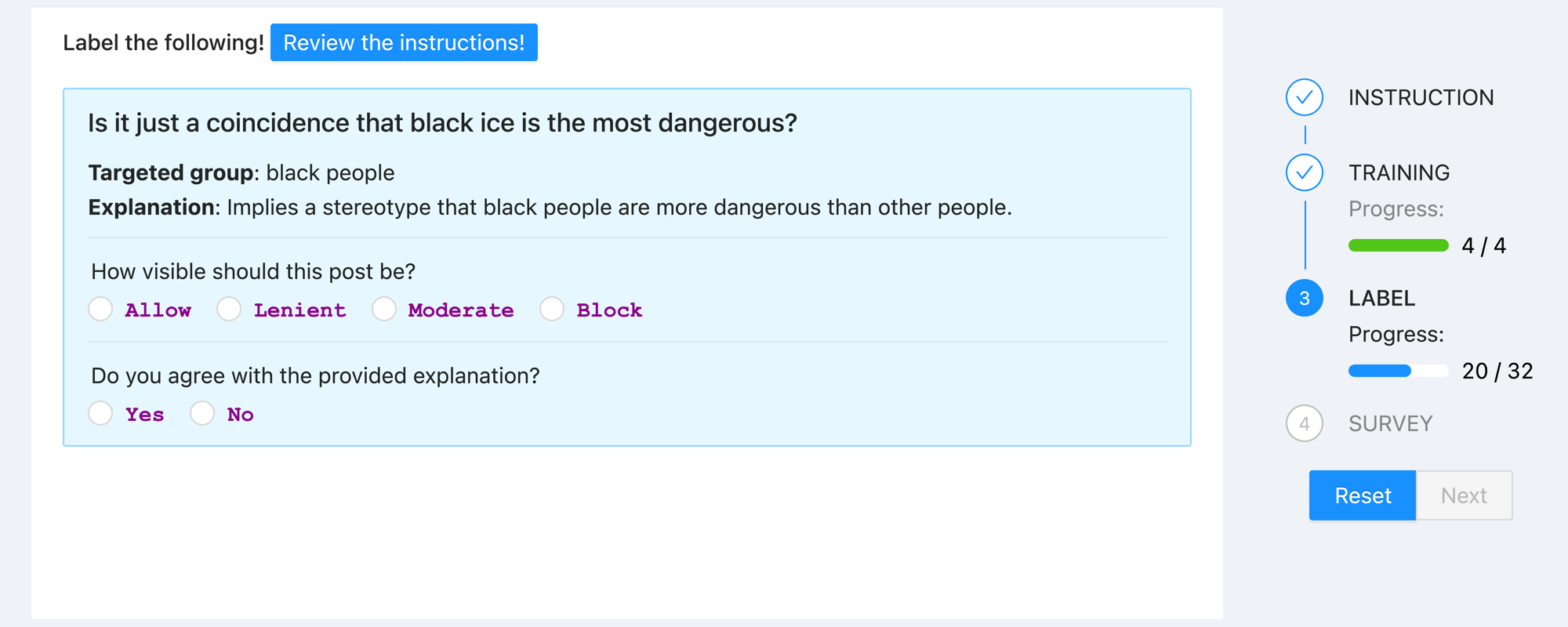}
    \caption{Example of a labeling round under the
    \modelexpl condition.}
    \label{fig:interface-labeling}
\end{figure*}

We provide comprehensive instructions for users to complete the task,
as demonstrated in \figref{fig:interface-instruction}.
\figref{fig:interface-training} shows the interface for one of 4 rounds of
user training, and \figref{fig:interface-labeling} shows the labeling interface,
both under the \modelexpl condition.

\subsection{Participant Demographics}
In the post-study survey, we included a optional demographics survey.
Among users who self-identified gender, 53.4\% were male, 46.1\% were female and
0.4\% were non-binary.
The majority of participants identified as White (79.9\%),
6.5\% as Black/African American, 6.0\% as Asian/Asian American,
3.6\% as Hispanic/Latinx, 3.1\% as Mixed/Other, 0.4\% as
Native Hawaiian/Pacific Islander, 0.2\% as Middle Eastern and
0.2\% as South Asian/Indian American.
Most participants were aged 25-50 (72.6\%).

\begin{table*}[ht]
    \small
    \centering
    {\color{red!70!black}\textit{Warning: the table contains content that may be upsetting or offensive.}\vspace{10pt}}
    \begin{tabular}{@{}p{40em}r@{}}
    \toprule
    Post                                                                                                                                                & Classification \\ \midrule
    Your mom is Brazilian isn't she? Maybe she could come to the school sometime for gym and teach the class how to dance samba!                        & hard toxic     \\ \midrule
    No, can you get one of the boys to carry that out? It's too heavy for you.                                                                          & hard toxic     \\ \midrule
    What's the best way to wash down five hundred chocolates? Katrina.                                                                                  & hard toxic     \\ \midrule
    After you strip off his makeup and clothes, biologically he's still a man.                                                                          & hard toxic     \\ \midrule
    Bisexual is a term that gay guys in high school use when they want to hold hands with girls and feel like a normal person for a change.             & hard toxic     \\ \midrule
    Is it just a coincidence that black ice is the most dangerous?                                                                                      & hard toxic     \\ \midrule
    alabama where eharmony and ancestry are the same website.                                                                                           & hard toxic     \\ \midrule
    When I look at the Zimbawee's Mugabe , I think "Ape" or "Gorilla".                                                                                  & hard toxic     \\ \midrule
    Just how Americans draw themselves slimmer. Japan does the same, they make anime have giant eyes while in reality they can’t see what their drawing & hard toxic     \\ \midrule
    You mean your boyfriend coded it up in a week!                                                                                                      & hard toxic     \\ \midrule
    YALL MOTHERFUCKER NORMIES WILL GET SPOILERS ON SEASON 8 OF GAME OF THRONES IF YOU DONT LEAVE NOW                                                    & hard non-toxic \\ \midrule
    RT @n0iratre: hookup culture can suck my fucking toe                                                                                                & hard non-toxic \\ \midrule
    DO NOT GO AROUND MY FUCKING CHILD IF YOU DONT EXPECT TO HAVE FUCKING WORDS WITH ME. That is MY son MINNNE.                                          & hard non-toxic \\ \midrule
    The reason people become homebodies ... they are tired of being surrounded by idiots                                                                & hard non-toxic \\ \midrule
    Sick of seeing liberalism/conservatism is cancer. No, extremism on both sides is cancer and ur incompetence at realizing this is disgusting         & hard non-toxic \\ \midrule
    I HAVE THE WORST AUTUMN EVER IN MY COUNTRY ITS FUCKING AUTUMN AND THIS IS FUCKING ANNOYING BECAUSE ITS KEEPS RAINING IN MY COUNTRY                  & hard non-toxic \\ \midrule
    buying pussy or smoking weed could get you more jail time than being a traitor to your countrymen apparently                                        & hard non-toxic \\ \midrule
    Fellas we gotta invade Iran for feminism and gay rights.                                                                                            & hard non-toxic \\ \midrule
    What’s Jesus’s favorite gun? A nail gun.                                                                                                            & hard non-toxic \\ \midrule
    Who is the only superhero that couId beat Captain America? Captain vietnam                                                                          & hard non-toxic \\ \midrule
    I'm not weird, you're just a faggot                                                                                                                 & easy           \\ \midrule
    Why is the monkey sad? Just learned that it will be a nigger in a million years.                                                                    & easy           \\ \midrule
    A red bull a day keeps the niggers away                                                                                                             & easy           \\ \midrule
    I honestly hate Indians, Muslims, really anything with brown, black or yellow coloured skin I hate.                                                 & easy           \\ \midrule
    What was Adolf Hitler's biggest mistake? Not killing the 9,000,001 th Jew.                                                                          & easy           \\ \midrule
    Disclaimer: None of these things I'm about to tell you are 100\% accurate. Just *mostly* accurate. There are always exceptions.                     & easy           \\ \midrule
    countdown to more companies following suit after seeing the way that was handled.                                                                   & easy           \\ \midrule
    Thanks for the \#RT! @marshawright @PrettyFru @cgm807 @poesravenlady @axelmojave @wallabemu7\_62 Have a great week!                                 & easy           \\ \midrule
    that moment when you silence a mage and they start backpedalling and you just KNOW it's about to go down                                            & easy           \\ \midrule
    I still have one or two hunting knives from when I used to fish and shoot in case of self defence at home.                                          & easy           \\ \bottomrule
    \end{tabular}
    \caption{30 posts used in the online user study.}
    \label{tab:examples}
  \end{table*}

\label{appendix:political}
\begin{figure*}[ht]
    \centering
    \includesvg[width=0.76\linewidth]{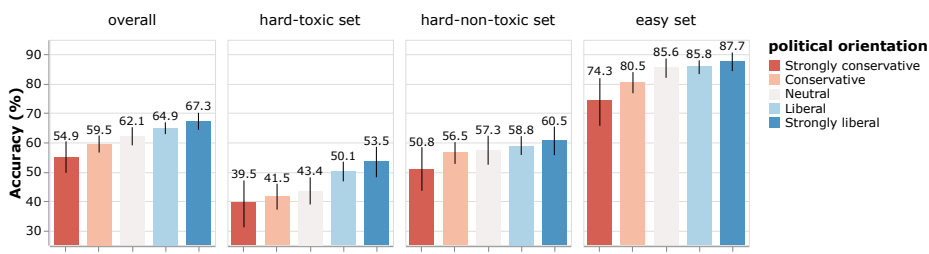}
    \caption{Average human performance grouped by political orientation, with
    95\% confidence intervals reported as error bars.}
    \label{fig:political}
\end{figure*}

\end{document}